\documentclass[conference]{IEEEtran}
\usepackage{cite}
\usepackage{amsmath,amssymb,amsfonts}
\usepackage{algorithmic}
\usepackage{graphicx}
\usepackage{textcomp}
\usepackage{xcolor}
\usepackage{balance}
\usepackage{fancyhdr}
\usepackage{url}
\usepackage{hyperref}
\usepackage[square,comma,sort&compress,numbers]{natbib} 
\usepackage{threeparttable,multirow,adjustbox,threeparttable}
\usepackage{orcidlink} 
\usepackage{adjustbox}
\usepackage{verbatim}
\usepackage{hyperref}       
\hypersetup{
    colorlinks=true,        
    linkcolor=blue,         
    citecolor=blue,         
    filecolor=blue,         
    urlcolor=blue           
}
\usepackage{caption}
\renewcommand{\arraystretch}{1.3}
\begin{document}
\title{Masked adversarial neural network for cell type deconvolution in spatial transcriptomics}
\author{
\IEEEauthorblockN{
Lin Huang$^\dag$, 
Xiaofei Liu$^\dag$, 
Shunfang Wang,
and Wenwen Min$^{*}\orcidlink{0000-0002-2558-2911} $}\\
\IEEEauthorblockA{
School of Information Science and Engineering, Yunnan University, Kunming 650091, Yunnan, China\\
\dag Co-first author. 
$^{*}$Correspondence author: 
\href{email}{minwenwen@ynu.edu.cn}
}			
}
\maketitle

\thispagestyle{fancy}
\begin{abstract}
Accurately determining cell type composition in disease-relevant tissues is crucial for identifying disease targets. Most existing spatial transcriptomics (ST) technologies cannot achieve single-cell resolution, making it challenging to accurately determine cell types. To address this issue, various deconvolution methods have been developed. Most of these methods use single-cell RNA sequencing (scRNA-seq) data from the same tissue as a reference to infer cell types in ST data spots. However, they often overlook the differences between scRNA-seq and ST data. To overcome this limitation, we propose a Masked Adversarial Neural Network (MACD). MACD employs adversarial learning to align real ST data with simulated ST data generated from scRNA-seq data. By mapping them into a unified latent space, it can minimize the differences between the two types of data. Additionally, MACD uses masking techniques to effectively learn the features of real ST data and mitigate noise. We evaluated MACD on 32 simulated datasets and 2 real datasets, demonstrating its accuracy in performing cell type deconvolution.
All code and public datasets used in this paper are available at \url{https://github.com/wenwenmin/MACD} and \url{https://zenodo.org/records/12804822}.

\end{abstract}


\section{Introduction}
Spatial transcriptomics (ST) data integrate gene expression profiles with spatial information, providing crucial insights for disease research and targeted therapeutic strategies \cite{jain2024spatial,rao2021exploring,Min2024a}. 
ST data for cancer research enables the detailed mapping of gene expression patterns within tumors, enhancing our understanding of tumor heterogeneity, microenvironment interactions, and the identification of potential biomarkers \cite{wang2023spatially,min2024dimensionality}.

Currently, ST technologies are classified into image-based and sequencing-based methods \cite{bressan2023dawn}. Image-based methods, like MERFISH \cite{MERFISH} and seqFISH+ \cite{seqFISH}, offer high spatial resolution but typically detect only a few hundred genes \cite{Moffitt2018,li2024spadit}. In contrast, sequencing-based methods, such as 10X Genomics Visium \cite{10X_Visium}, are more widely used because they can detect a broader range of genes. Nevertheless, these methods generally exhibit lower spatial resolution, with each spot potentially containing multiple cells \cite{Miller2021}. 

To address these limitations, cell type deconvolution methods have been developed to disentangle cellular composition within ST data, mitigating the constraints of single-cell resolution. Among existing cell type deconvolution methods, traditional approaches such as CIBERSORT \cite{Newman2015} and SPOTlight \cite{SPOTlight} often directly utilize single-cell RNA sequencing (scRNA-seq) data as references to infer cellular compositions in ST data. 
CIBERSORT, initially designed for bulk RNA-seq data, uses linear support vector regression, assuming scRNA-seq profiles represent aggregated ST signals. Similarly, SPOTlight employs non-negative matrix factorization to decompose ST data, assuming scRNA-seq data are suitable references.
However, these methods often overlook critical differences: scRNA-seq provides gene expression profiles from individual cells, whereas ST data represent aggregated gene expression from multiple cells \cite{Stuart2019}. Moreover, variations in sample preparation and technical noise introduce additional discrepancies \cite{Rybakov2020}. These factors may make the deconvolution problem more difficult and the results potentially less accurate. 
DestVI \cite{DestVI} integrates variational inference with deep learning techniques to estimate cellular composition at each spatial location within tissue samples. Yet, its probabilistic modeling may not fully address the inherent discrepancies between ST and scRNA-seq data. To mitigate these discrepancies, spoint \cite{spoint} improves alignment by generating simulated ST data and utilizing the maximum mean discrepancy loss function. However, spoint does not specifically target the unique characteristics of real ST data during feature extraction, which may limit its effectiveness.
\begin{figure*}[h]
\centering
\includegraphics[width=1\textwidth]{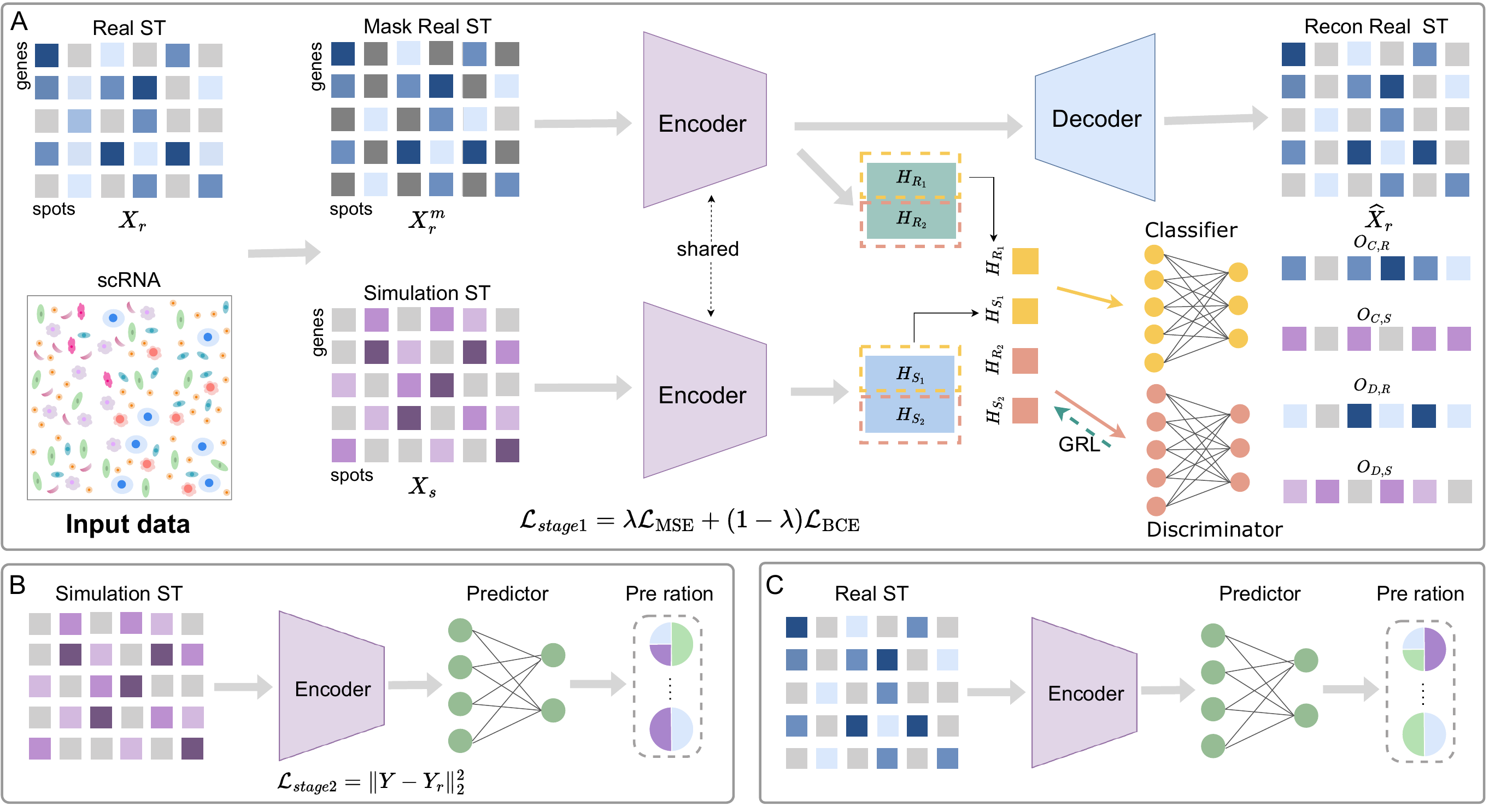}
    \caption{The network architecture of MACD. The MACD training phase consists of two stages (\textbf{A}) and (\textbf{B}). 
   The simulated and masked ST data are processed by a shared \textbf{Encoder} to produce latent variables using masked autoencoder. 
    Adversarial learning is performed on these latent variables, where a classifier distinguishes between real and simulated ST data, and a discriminator, utilizing a Gradient Reversal Layer (GRL), is trained to obscure the differences between them. 
    (\textbf{C}) The trained model is used to infer cell type of the real ST data.}
\label{fig1}
\end{figure*} 

To this end, we propose MACD, a masked adversarial neural network for cell type deconvolution in spatial transcriptomics. 
MACD first employs masked mechanism, i.e., a masked autoencoder \cite{he2022masked} uncovers the intrinsic characteristics of real ST data.
Additionally, MACD employs an adversarial neural network \cite{wang2023deep} comprising a classifier and a discriminator to optimize the data encoding process. This adversarial setup makes it challenging for the model to distinguish between real and simulated ST data after encoding, thereby minimizing discrepancies and enhancing the consistency and comparability of the data. MACD utilizes labeled simulated ST data to guide model training, enabling accurate inference of cellular composition in simulated ST data. Since the model becomes proficient at distinguishing between real and simulated ST data, it is also expected to perform well in accurately predicting the cellular composition of real ST data. 

The main contributions of the paper are as follows:
\begin{itemize}
    \item A novel adversarial learning-based MACD method is proposed, which aligns real ST data with simulated ST data in the latent space. This approach significantly reduces data discrepancies and enhances consistency, thereby improving the accuracy of cell type deconvolution.
   \item Integrating a masked autoencoder into MACD uncovers the intrinsic characteristics of real ST data, enhancing the model's understanding of data nuances.
   \item Integrating adversarial learning into MACD removes batch effects between real ST and simulated ST data, enabling the supervised learning model built in the latent space of simulated ST data to be effectively applied for cell type deconvolution in real ST data.
   \item Extensive validation of MACD with 32 simulated datasets and 2 real datasets demonstrates its exceptional robustness and performance across various conditions.
\end{itemize}

\begin{table*}[htbp]%
\centering
\caption{Detailed information of each  dataset. The scRNA-seq data in Dataset1 to Dataset32 are real data, and the matching ST data are simulated, while the scRNA-seq and ST data in MLN and HDH are both real data.}\label{tab1}
\begin{adjustbox}{width=1\textwidth}
\renewcommand{\arraystretch} {0.91} 
\fontsize{9}{10}\selectfont  
\begin{tabular}{l|l|cccc|cccccc}
\hline
\multirow{3}{*}{\textbf{Datasets}} & \multirow{3}{*}{\textbf{Tissue}} & \multicolumn{4}{c|}{\textbf{source\_data}}                                                                        & \multicolumn{6}{c}{\textbf{preproce\_data}}                                                                                                                            \\ \cline{3-12} 
                                   &                                  & \multicolumn{2}{l|}{\textbf{Number of Cells/Spots}}                & \multicolumn{2}{c|}{\textbf{Number of Genes}} & \multicolumn{2}{l|}{\textbf{Number of Cells/Spots}}                & \multicolumn{2}{l|}{\textbf{Number of Genes}}          & \multicolumn{2}{l}{\textbf{Dropout rate}} \\  \cline{3-12} 
                                   &                                  & \multicolumn{1}{c|}{SC (Cells)} & \multicolumn{1}{c|}{ST (Spots)} & \multicolumn{1}{c|}{SC}          & ST         & \multicolumn{1}{c|}{SC (Cells)} & \multicolumn{1}{c|}{ST (Spots)} & \multicolumn{1}{c|}{SC}   & \multicolumn{1}{c|}{ST}   & \multicolumn{1}{c|}{SC}        & ST        \\ \hline
Dataset1                           & Human-Brain                      & \multicolumn{1}{c|}{10000}      & \multicolumn{1}{c|}{1000}       & \multicolumn{1}{c|}{33691}       & 46732      & \multicolumn{1}{c|}{10000}      & \multicolumn{1}{c|}{1000}       & \multicolumn{1}{c|}{4456} & \multicolumn{1}{c|}{4456} & \multicolumn{1}{c|}{81.29\%}   & 61.01\%   \\
Dataset2                           & Mouse-Brain                      & \multicolumn{1}{c|}{10000}      & \multicolumn{1}{c|}{1000}       & \multicolumn{1}{c|}{43324}       & 24683      & \multicolumn{1}{c|}{10000}      & \multicolumn{1}{c|}{1000}       & \multicolumn{1}{c|}{4481} & \multicolumn{1}{c|}{4481} & \multicolumn{1}{c|}{61.05\%}   & 76.30\%   \\
Dataset3                           & Human-Liver                      & \multicolumn{1}{c|}{3821}       & \multicolumn{1}{c|}{1000}       & \multicolumn{1}{c|}{18328}       & 19850      & \multicolumn{1}{c|}{3821}       & \multicolumn{1}{c|}{1000}       & \multicolumn{1}{c|}{1186} & \multicolumn{1}{c|}{1186} & \multicolumn{1}{c|}{90.06\%}   & 94.83\%   \\
Dataset4                           & Human-Liver                      & \multicolumn{1}{c|}{6948}       & \multicolumn{1}{c|}{1000}       & \multicolumn{1}{c|}{20007}       & 26160      & \multicolumn{1}{c|}{6948}       & \multicolumn{1}{c|}{1000}       & \multicolumn{1}{c|}{1171} & \multicolumn{1}{c|}{1171} & \multicolumn{1}{c|}{92.08\%}   & 86.78\%   \\
Dataset5                           & Human-Lung                       & \multicolumn{1}{c|}{10000}      & \multicolumn{1}{c|}{1000}       & \multicolumn{1}{c|}{25734}       & 38150      & \multicolumn{1}{c|}{10000}      & \multicolumn{1}{c|}{1000}       & \multicolumn{1}{c|}{2303} & \multicolumn{1}{c|}{2303} & \multicolumn{1}{c|}{88.12\%}   & 95.32\%   \\
Dataset6                           & Human-Lung                       & \multicolumn{1}{c|}{10000}      & \multicolumn{1}{c|}{1000}       & \multicolumn{1}{c|}{25734}       & 26828      & \multicolumn{1}{c|}{10000}      & \multicolumn{1}{c|}{1000}       & \multicolumn{1}{c|}{1639} & \multicolumn{1}{c|}{1639} & \multicolumn{1}{c|}{84.16\%}   & 89.17\%   \\
Dataset7                           & Human-Lung                       & \multicolumn{1}{c|}{10000}      & \multicolumn{1}{c|}{1000}       & \multicolumn{1}{c|}{22066}       & 25199      & \multicolumn{1}{c|}{10000}      & \multicolumn{1}{c|}{1000}       & \multicolumn{1}{c|}{2831} & \multicolumn{1}{c|}{2831} & \multicolumn{1}{c|}{88.54\%}   & 90.01\%   \\
Dataset8                           & Human-Kidney                     & \multicolumn{1}{c|}{10000}      & \multicolumn{1}{c|}{1000}       & \multicolumn{1}{c|}{27345}       & 31489      & \multicolumn{1}{c|}{10000}      & \multicolumn{1}{c|}{1000}       & \multicolumn{1}{c|}{1531} & \multicolumn{1}{c|}{1531} & \multicolumn{1}{c|}{90.28\%}   & 95.15\%   \\
Dataset9                           & Mouse-Kidney                     & \multicolumn{1}{c|}{10000}      & \multicolumn{1}{c|}{1000}       & \multicolumn{1}{c|}{24965}       & 29244      & \multicolumn{1}{c|}{10000}      & \multicolumn{1}{c|}{1000}       & \multicolumn{1}{c|}{1120} & \multicolumn{1}{c|}{1120} & \multicolumn{1}{c|}{96.76\%}   & 95.51\%   \\
Dataset10                          & Human-Heart                      & \multicolumn{1}{c|}{10000}      & \multicolumn{1}{c|}{1000}       & \multicolumn{1}{c|}{17926}       & 29484      & \multicolumn{1}{c|}{10000}      & \multicolumn{1}{c|}{1000}       & \multicolumn{1}{c|}{1426} & \multicolumn{1}{c|}{1426} & \multicolumn{1}{c|}{82.13\%}   & 95.66\%   \\
Dataset11                          & Human-Heart                      & \multicolumn{1}{c|}{10000}      & \multicolumn{1}{c|}{1000}       & \multicolumn{1}{c|}{17926}       & 31580      & \multicolumn{1}{c|}{10000}      & \multicolumn{1}{c|}{1000}       & \multicolumn{1}{c|}{1426} & \multicolumn{1}{c|}{1426} & \multicolumn{1}{c|}{82.75\%}   & 92.18\%   \\
Dataset12                          & Human-Pancreas                   & \multicolumn{1}{c|}{2282}       & \multicolumn{1}{c|}{1000}       & \multicolumn{1}{c|}{21198}       & 17499      & \multicolumn{1}{c|}{2282}       & \multicolumn{1}{c|}{1000}       & \multicolumn{1}{c|}{1077} & \multicolumn{1}{c|}{1077} & \multicolumn{1}{c|}{79.84\%}   & 85.17\%   \\
Dataset13                          & Human-Pancreas                   & \multicolumn{1}{c|}{1040}       & \multicolumn{1}{c|}{1000}       & \multicolumn{1}{c|}{21625}       & 17499      & \multicolumn{1}{c|}{1040}       & \multicolumn{1}{c|}{1000}       & \multicolumn{1}{c|}{1427} & \multicolumn{1}{c|}{1427} & \multicolumn{1}{c|}{64.85\%}   & 85.48\%   \\
Dataset14                          & Human-Pancreas                   & \multicolumn{1}{c|}{943}        & \multicolumn{1}{c|}{1000}       & \multicolumn{1}{c|}{21625}       & 21198      & \multicolumn{1}{c|}{943}        & \multicolumn{1}{c|}{1000}       & \multicolumn{1}{c|}{1135} & \multicolumn{1}{c|}{1135} & \multicolumn{1}{c|}{66.08\%}   & 45.05\%   \\
Dataset15                          & Mouse-Pancreas                   & \multicolumn{1}{c|}{1382}       & \multicolumn{1}{c|}{1000}       & \multicolumn{1}{c|}{19745}       & 14860      & \multicolumn{1}{c|}{1382}       & \multicolumn{1}{c|}{1000}       & \multicolumn{1}{c|}{1504} & \multicolumn{1}{c|}{1504} & \multicolumn{1}{c|}{67.44\%}   & 84.52\%   \\
Dataset16                          & Mouse-Trachea                    & \multicolumn{1}{c|}{6937}       & \multicolumn{1}{c|}{1000}       & \multicolumn{1}{c|}{27084}       & 18388      & \multicolumn{1}{c|}{6937}       & \multicolumn{1}{c|}{1000}       & \multicolumn{1}{c|}{1034} & \multicolumn{1}{c|}{1034} & \multicolumn{1}{c|}{87.83\%}   & 87.74\%   \\
Dataset17                          & Human-Brain                      & \multicolumn{1}{c|}{10000}      & \multicolumn{1}{c|}{1000}       & \multicolumn{1}{c|}{46732}       & 33691      & \multicolumn{1}{c|}{10000}      & \multicolumn{1}{c|}{1000}       & \multicolumn{1}{c|}{4942} & \multicolumn{1}{c|}{4942} & \multicolumn{1}{c|}{85.99\%}   & 79.82\%   \\
Dataset18                          & Mouse-Brain                      & \multicolumn{1}{c|}{9999}       & \multicolumn{1}{c|}{1000}       & \multicolumn{1}{c|}{24683}       & 43324      & \multicolumn{1}{c|}{9999}       & \multicolumn{1}{c|}{1000}       & \multicolumn{1}{c|}{3227} & \multicolumn{1}{c|}{3227} & \multicolumn{1}{c|}{78.03\%}   & 37.39\%   \\
Dataset19                          & Human-Liver                      & \multicolumn{1}{c|}{10000}      & \multicolumn{1}{c|}{1000}       & \multicolumn{1}{c|}{19850}       & 18328      & \multicolumn{1}{c|}{10000}      & \multicolumn{1}{c|}{1000}       & \multicolumn{1}{c|}{1183} & \multicolumn{1}{c|}{1183} & \multicolumn{1}{c|}{95.88\%}   & 87.03\%   \\
Dataset20                          & Human-Liver                      & \multicolumn{1}{c|}{8785}       & \multicolumn{1}{c|}{1000}       & \multicolumn{1}{c|}{26160}       & 20007      & \multicolumn{1}{c|}{8785}       & \multicolumn{1}{c|}{1000}       & \multicolumn{1}{c|}{1101} & \multicolumn{1}{c|}{1101} & \multicolumn{1}{c|}{90.33\%}   & 90.97\%   \\
Dataset21                          & Human-Lung                       & \multicolumn{1}{c|}{10000}      & \multicolumn{1}{c|}{1000}       & \multicolumn{1}{c|}{38150}       & 25734      & \multicolumn{1}{c|}{10000}      & \multicolumn{1}{c|}{1000}       & \multicolumn{1}{c|}{1813} & \multicolumn{1}{c|}{1813} & \multicolumn{1}{c|}{96.07\%}   & 55.29\%   \\
Dataset22                          & Human-Lung                       & \multicolumn{1}{c|}{10000}      & \multicolumn{1}{c|}{1000}       & \multicolumn{1}{c|}{26828}       & 25734      & \multicolumn{1}{c|}{10000}      & \multicolumn{1}{c|}{1000}       & \multicolumn{1}{c|}{1502} & \multicolumn{1}{c|}{1502} & \multicolumn{1}{c|}{90.63\%}   & 44.66\%   \\
Dataset23                          & Human-Lung                       & \multicolumn{1}{c|}{10000}      & \multicolumn{1}{c|}{1000}       & \multicolumn{1}{c|}{25199}       & 22066      & \multicolumn{1}{c|}{10000}      & \multicolumn{1}{c|}{1000}       & \multicolumn{1}{c|}{2535} & \multicolumn{1}{c|}{2535} & \multicolumn{1}{c|}{92.00\%}   & 86.07\%   \\
Dataset24                          & Human-Kidney                     & \multicolumn{1}{c|}{10000}      & \multicolumn{1}{c|}{1000}       & \multicolumn{1}{c|}{31489}       & 27345      & \multicolumn{1}{c|}{10000}      & \multicolumn{1}{c|}{1000}       & \multicolumn{1}{c|}{1218} & \multicolumn{1}{c|}{1218} & \multicolumn{1}{c|}{95.91\%}   & 87.56\%   \\
Dataset25                          & Mouse-Kidney                     & \multicolumn{1}{c|}{10000}      & \multicolumn{1}{c|}{1000}       & \multicolumn{1}{c|}{29244}       & 24965      & \multicolumn{1}{c|}{10000}      & \multicolumn{1}{c|}{1000}       & \multicolumn{1}{c|}{885}  & \multicolumn{1}{c|}{885}  & \multicolumn{1}{c|}{96.63\%}   & 94.99\%   \\
Dataset26                          & Human-Heart                      & \multicolumn{1}{c|}{10000}      & \multicolumn{1}{c|}{1000}       & \multicolumn{1}{c|}{29484}       & 17926      & \multicolumn{1}{c|}{10000}      & \multicolumn{1}{c|}{1000}       & \multicolumn{1}{c|}{1077} & \multicolumn{1}{c|}{1077} & \multicolumn{1}{c|}{95.42\%}   & 75.15\%   \\
Dataset27                          & Human-Heart                      & \multicolumn{1}{c|}{10000}      & \multicolumn{1}{c|}{1000}       & \multicolumn{1}{c|}{31580}       & 17926      & \multicolumn{1}{c|}{10000}      & \multicolumn{1}{c|}{1000}       & \multicolumn{1}{c|}{1358} & \multicolumn{1}{c|}{1358} & \multicolumn{1}{c|}{93.58\%}   & 77.14\%   \\
Dataset28                          & Human-Pancreas                   & \multicolumn{1}{c|}{7944}       & \multicolumn{1}{c|}{1000}       & \multicolumn{1}{c|}{17499}       & 21198      & \multicolumn{1}{c|}{7944}       & \multicolumn{1}{c|}{1000}       & \multicolumn{1}{c|}{1148} & \multicolumn{1}{c|}{1148} & \multicolumn{1}{c|}{87.47\%}   & 41.59\%   \\
Dataset29                          & Human-Pancreas                   & \multicolumn{1}{c|}{8494}       & \multicolumn{1}{c|}{1000}       & \multicolumn{1}{c|}{17499}       & 21625      & \multicolumn{1}{c|}{8494}       & \multicolumn{1}{c|}{1000}       & \multicolumn{1}{c|}{1559} & \multicolumn{1}{c|}{1559} & \multicolumn{1}{c|}{87.80\%}   & 32.99\%   \\
Dataset30                          & Human-Pancreas                   & \multicolumn{1}{c|}{2282}       & \multicolumn{1}{c|}{1000}       & \multicolumn{1}{c|}{21198}       & 21625      & \multicolumn{1}{c|}{2282}       & \multicolumn{1}{c|}{1000}       & \multicolumn{1}{c|}{1080} & \multicolumn{1}{c|}{1080} & \multicolumn{1}{c|}{81.47\%}   & 38.31\%   \\
Dataset31                          & Mouse-Pancreas                   & \multicolumn{1}{c|}{1827}       & \multicolumn{1}{c|}{1000}       & \multicolumn{1}{c|}{14860}       & 19745      & \multicolumn{1}{c|}{1827}       & \multicolumn{1}{c|}{1000}       & \multicolumn{1}{c|}{1143} & \multicolumn{1}{c|}{1143} & \multicolumn{1}{c|}{87.02\%}   & 24.93\%   \\
Dataset32                          & Mouse-Trachea                    & \multicolumn{1}{c|}{7128}       & \multicolumn{1}{c|}{1000}       & \multicolumn{1}{c|}{18388}       & 27084      & \multicolumn{1}{c|}{7128}       & \multicolumn{1}{c|}{1000}       & \multicolumn{1}{c|}{1006} & \multicolumn{1}{c|}{1006} & \multicolumn{1}{c|}{88.17\%}   & 82.86\%   \\
\hline
MLN                         & Murine lymph node                & \multicolumn{1}{c|}{14989}      & \multicolumn{1}{c|}{1092}       & \multicolumn{1}{c|}{12854}       & 13948      & \multicolumn{1}{c|}{14989}      & \multicolumn{1}{c|}{1092}       & \multicolumn{1}{c|}{1870} & \multicolumn{1}{c|}{1870} & \multicolumn{1}{c|}{83.41\%}   & 61.09\%   \\
HDH                          & Human developing heart                      & \multicolumn{1}{c|}{3777}       & \multicolumn{1}{c|}{210}        & \multicolumn{1}{c|}{15323}       & 38936      & \multicolumn{1}{c|}{3777}       & \multicolumn{1}{c|}{209}        & \multicolumn{1}{c|}{2373} & \multicolumn{1}{c|}{2373} & \multicolumn{1}{c|}{81.00\%}   & 94.01\%   \\ \hline
\end{tabular}
\end{adjustbox}
\end{table*}

\section{PROPOSED METHODS}\label{MATERIALS AND MRTHODS}
\subsection{Overview of the proposed MACD method}\label{Multimodal framework}
Herein, a masked adversarial neural network (MACD) is proposed for cell type deconvolution of ST data (\autoref{fig1}). During training, MACD first employs a masked autoencoder to capture the features of the real ST data. Then, adversarial learning aligns real and simulated ST data in the latent space, thereby eliminating discrepancies between them. Finally, supervised learning is conducted on labeled simulated ST data to infer the cell type composition in spatial spots. 
\begin{figure*}[h]%
    \centering
    \includegraphics[width=1\textwidth]{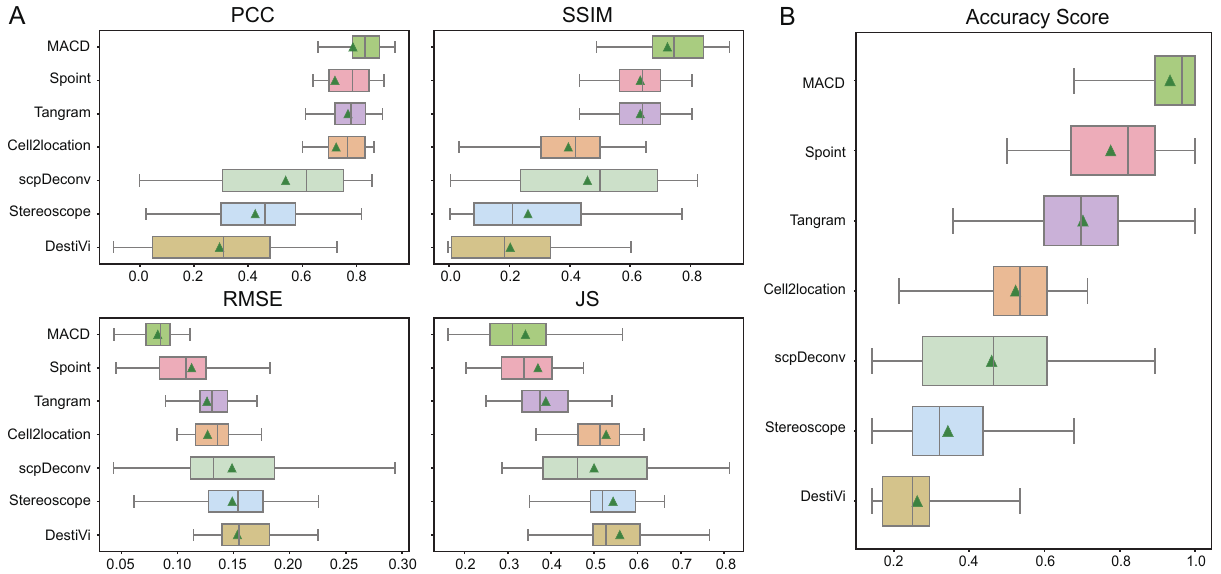}
    \caption{
    Performance evaluation of MACD on 32 simulated datasets, with higher PCC, SSIM, and AS values, and lower RMSE and JS values indicating better performance. AS (Accuracy Score) is a composite metric that combines PCC, SSIM, RMSE, and JS. Green triangles represent the mean values, and the middle line represents the median.
    }\label{fig2}
\end{figure*}
\begin{figure*}[htbp]%
\centering
\includegraphics[width=.96\textwidth]{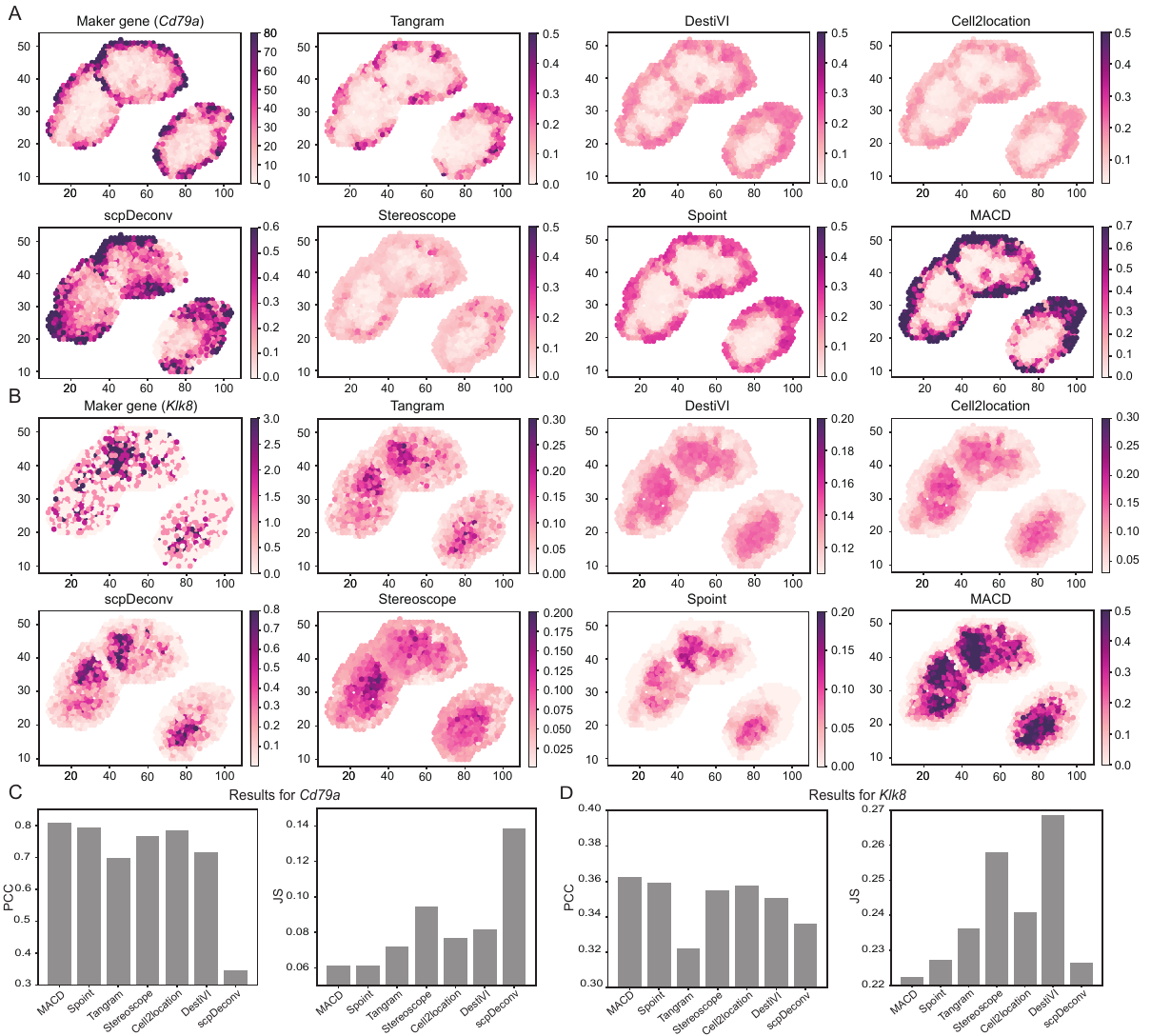}
\caption{MACD effectively analyzes both major and rare cell types in the Murine Lymph Node (MLN) dataset.
(\textbf{A}) The first panel displays the expression levels of the marker gene \textit{Cd79a}, followed by the proportions of Mature B cells (a rare cell type) estimated by MACD and other methods.
(\textbf{B}) The first panel shows the expression levels of the marker gene \textit{Klk8}, followed by the proportions of CD8 T cells (a major cell type) estimated by MACD and other methods.
(\textbf{C}) Compares the PCC and JS between MACD and other methods for the marker gene \textit{Cd79a} and the proportion of Mature B cells.
(\textbf{D}) Compares the PCC and JS between MACD and other methods for the marker gene \textit{Klk8} and the proportion of CD8 T cells.
}\label{fig3}
\end{figure*}
\begin{figure*}[htbp]%
\centering
\includegraphics[width=0.96\textwidth]{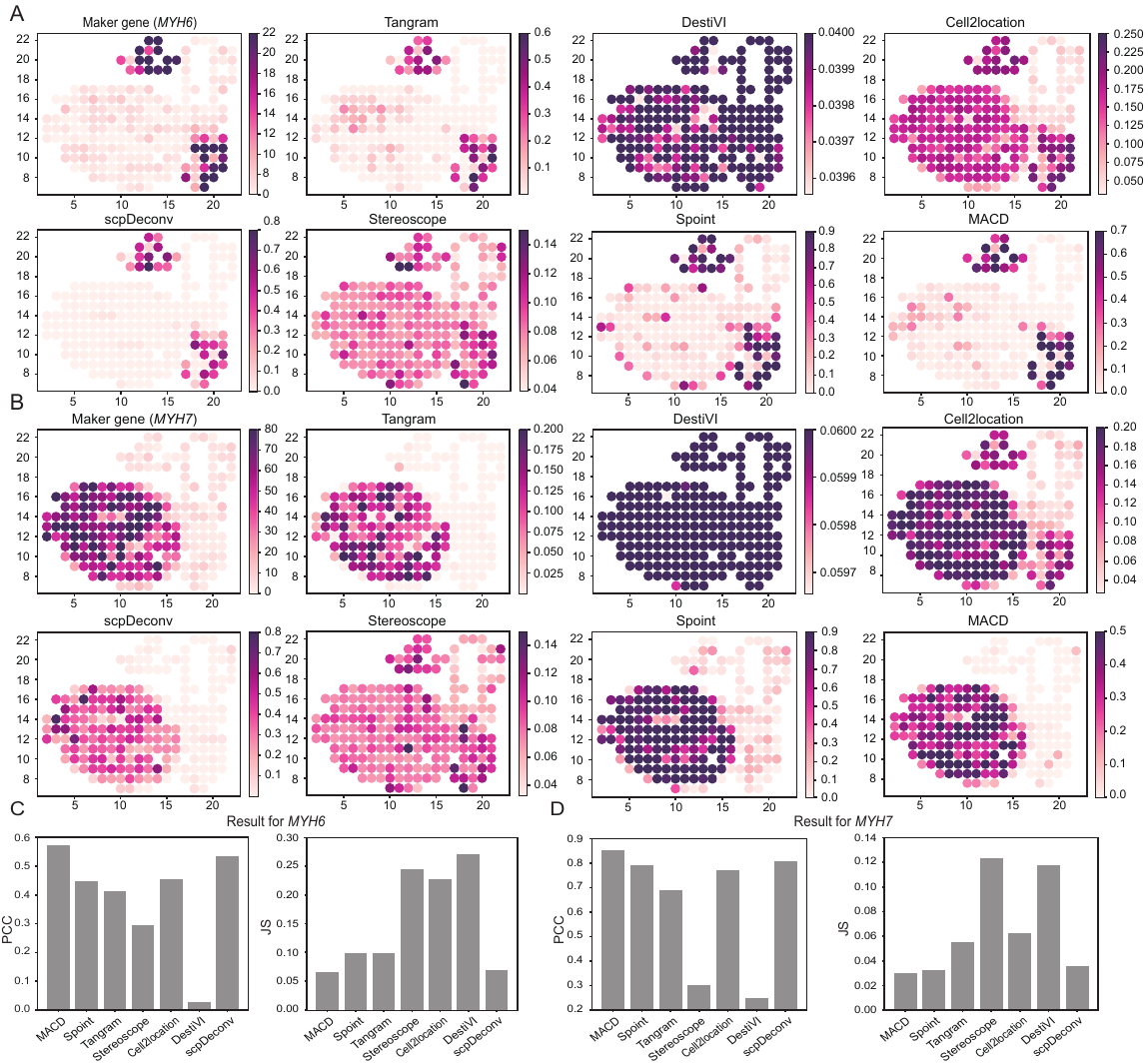}
\caption{MACD effectively analyzes both major and rare cell types in the Human Developing Heart (HDH) dataset.
(\textbf{A}) The first panel shows the expression levels of the marker gene \textit{MYH6} , followed by the proportions of Atrial cardiomyocytes (a rare cell type) estimated by MACD and other methods.
(\textbf{B}) The first panel displays the expression levels of the marker gene \textit{MYH7}, followed by the proportions of Ventricular cardiomyocytes (a major cell type) estimated by MACD and other methods.
(\textbf{C}) Compares the PCC and JS between MACD and other methods for the marker gene \textit{MYH6} and the proportion of Atrial cardiomyocytes.
(\textbf{D}) Compares the PCC and JS between MACD and other methods for the marker gene \textit{MYH7} and the proportion of Ventricular cardiomyocytes.}\label{fig4}
\end{figure*}
\subsection{Masked autoencoder learning}\label{MAL}
We apply a random masking strategy with a masking rate of \(\rho\) to the real ST data matrix ($X_r$) to obtain the new masked data matrix ($X_r^m$) (See~\autoref{fig1}A).

The \textbf{Encoder} \(\mathcal{E}(\cdot)\) processes the masked ST data through two stacked fully connected layers to generate a latent representation of gene expression:
\begin{equation}
\small{
    H_R = \mathcal{E}(X_r^m; \Theta_{e}) = W^1_e \mathrm{\phi}(\text{BN}(W^0_e X^m + b^0_e)) + b^1_e
}
\end{equation} 
where \(W^0_e\) and \(W^1_e\) are the weight matrices, \(b^0_e\) and \(b^1_e\) are the bias terms, \(\text{BN}\) denotes batch normalization, and \(\mathrm{\phi}\) represents the LeakyReLU activation function.

The \textbf{Decoder} \(\mathcal{D}(\cdot)\)  consists of three fully connected layers used to reconstruct the gene expression from the latent representation:
\begin{equation}
  {\widehat X_r} = \mathcal{D}(H_R;\Theta_{d})
    \label{eq-1}
\end{equation}
Similar to the Encoder, the Decoder employs fully connected layers with their respective weight matrices and bias terms, but it uses three layers instead of two. The detailed structure of the decoder is analogous to the encoder and therefore not expanded here.
\begin{equation}
 \mathcal{L}_{\text{MSE}} = \| (\widehat{X}_{r}- X_{r})\odot M \|_2^2
    \label{eq-8}
\end{equation}
where \(\odot\) denotes the element-wise multiplication. $M_{ij}$ is 1 or 0. If it is equal, it means that this element is Masked.

\subsection{Adversarial learning}\label{AL}
The encoder processes the masked real ST data and the simulated ST data, generating their corresponding latent representations. The first half of each latent representation is fed into a feedforward neural network (FNN)  \textbf{Classifier} \(\mathcal{F_C}(\cdot)\)  to produce predicted labels. The model is optimized to effectively distinguish between real and simulated data:
\begin{equation}
    O_{C,R} = \mathcal{F_C}(H_{R_1};\Theta_{C}),  O_{C,S} = \mathcal{F_C}(H_{S_1};\Theta_{C})
\end{equation}
where \(H_{R_1}\) and \(H_{S_1}\) are the first halves of the features from \(H_R\) and \(H_S\), respectively. \(H_S=\mathcal{E}(X_S;\Theta_{e})\), with \(X_S\) denoting the simulated ST data. \(\Theta_{C}\) represents the classifier's weights and biases. The loss of Classifier:
\begin{equation}\label{loss1}
       \mathcal{L}_{\text{C}} = f(O_{C,R}, O_{1,T}) + f(O_{C,S}, O_{0,T})
\end{equation}
where $O_{i,T} = [i, \ldots, i]\quad \text{for} \; i \in \{0, 1\}$, label \(i = 1\) indicates real ST data and \(i = 0\) indicates simulated ST data. And the binary cross entropy function is defined as
\begin{equation}\label{BCE}
f({y}_i,\hat{y}_i)= -\frac{1}{N} \sum_{i=1}^{N} \left[ y_i \log(\hat{y}_i) + (1 - y_i) \log(1 - \hat{y}_i) \right]
\end{equation}

Next, the remaining features are fed into a \textbf{Discriminator} \(\mathcal{F_D}(\cdot)\), implemented as  FNN. The discriminator incorporates a gradient reversal layer (GRL), which introduces a gradient reversal during training to prevent it from differentiating between real and simulated data in the latent space:
\begin{equation}
       O_{D,R} = \mathcal{F_D}(H_{R_2};\Theta_{D}), O_{D,S} = \mathcal{F_D}(H_{S_2};\Theta_{D})
    \label{eq-7}
\end{equation}
where \(H_{R2}\) and \(H_{S2}\) represent the remaining features of \(H_R\) and \(H_S\), respectively. \(\Theta_{D}\) represents the discriminator's weights and biases. The loss of Discriminator:
\begin{equation}
       \mathcal{L}_{\text{D}} = f(O_{D,R}, O_{1,T}) + f(O_{D,S}, O_{0,T})
    \label{loss2}
\end{equation}
Thus, the final loss function of the first stage of MACD (\autoref{fig1}A) is as follows:
\begin{equation}
\mathcal{L}_{stage1} =  \lambda \mathcal{L}_{\text{MSE}}+(1-\lambda)\mathcal{L}_{\text{BCE}} 
\label{allloss}
\end{equation}
where $\mathcal{L}_{\text{BCE}}= \mathcal{L}_{\text{C}} +\mathcal{L}_{\text{D}}$ (See \autoref{loss1} and \autoref{loss2}) and \(\lambda\) is used to balance the influences of reconstruction loss $\mathcal{L}_{\text{MSE}}$ and $\mathcal{L}_{\text{BCE}}$.

\subsection{Supervised prediction learning}\label{Supervised prediction training}
The second stage of MACD is a Supervised Prediction Learning module (See \autoref{fig1}B), which shares the \textbf{Encoder} with the first stage. The \textbf{Predictor} \(\mathcal{P}(\cdot)\) consists of two linear layers followed by a softmax layer, specifically designed to infer cell-type proportions from the latent representation:
\begin{equation}\label{eq-11}
\small
     Y = \mathcal{P}(H_S; \Theta_{P}) = \text{softmax}(W^1_p \mathrm{\phi}(\text{BN}(W^0_p H_S + b^0_p)) + b^1_p)
\end{equation}
where \(W^0_p\) and \(W^1_p\) are the weight matrices, \(b^0_p\) and \(b^1_p\) are the bias terms.

The loss function of the second stage of MACD is as follows:
\begin{equation}
    \mathcal{L}_{stage2}=\|Y-Y_r\|_2^2
\label{allloss1}
\end{equation}
where \(Y_r\) represents  the ground truth cell-type proportion of simulated ST data.

\subsection{Training process}\label{Training data}
The training procedure is divided into two phases. Initially, we use \autoref{allloss} as the training objective. In the subsequent phase, we shift to \autoref{allloss1} as the training objective. These two phases are alternated until the loss functions for both \autoref{allloss} and \autoref{allloss1} converge.

\subsection{Predicting process}\label{Prediction data}
The trained model applies to the real ST data to predict cell-type proportions:
\begin{equation}
   Y_t = \mathcal{P}(\mathcal{E}(X_r; \Theta_{e}); \Theta_{P}) 
    \label{eq-12}
\end{equation}
where \(Y_t\) represents the final predicted cell-type proportions of the real ST data.

\subsection{Evaluation metrics}\label{Evaluation metrics}
We use Pearson correlation coefficient (PCC), Jensen-Shannon divergence (JS), root mean square error (RMSE), Accuracy score (AS) and structural similarity index measure (SSIM) to evaluate the proposed method against baselines. 

\begin{equation}
    \text{PCC} = \frac{\text{cov}( x_i,\hat{x}_i)}{\sigma_i\hat{\sigma_i}}
    \label{eq-pcc}
\end{equation}
where \(x_i\) is the ground truth cell type composition for cell type \(i\), \(\sigma_i\) is its standard deviation, and \(\hat{x}_i\) and \(\hat{\sigma_i}\) are the predicted values.
\begin{equation}
    \text{SSIM} = \frac{(2\hat{u_i}u_i + C_1)(2\text{cov}( x_i,\hat{x}_i) + C_2)}{(\hat{u_i}^2 + u_i^2 + C_1)(\hat{\sigma_i}^2 +\sigma_i^2 + C_2)}
    \label{eq-ssim}
\end{equation}
where \(\mu_i\) is the average ground truth cell type composition for cell type \(i\), \(\hat{\mu}_i\) is the predicted average, and \(C_1\) and \(C_2\) are constants set to 0.01 and 0.03, respectively.
\begin{equation}
    \text{RMSE} = \sqrt{\frac{1}{M} \sum_{j=1}^M ( x_{ij} -\hat{x_{ij}})^2}
    \label{eq-rmse}
\end{equation}
where \(x_{ij}\) is the ground truth cell type composition for cell type \(i\) in spot \(j\), and \(\hat{x}_{ij}\) is the predicted value.
\begin{equation}
    \text{JS} =\frac{1}{2} \text{KL} \left( P_i, \frac{\hat{P_i} + P_i}{2} \right) + \frac{1}{2} \text{KL} \left( \hat{P_i}, \frac{\hat{P_i} + P_i}{2} \right) 
    \label{eq-js}
\end{equation}
where \(P_i\) and \(\hat{P}_i\) are the spatial distributions of cell type \(i\) in the ground truth and prediction, respectively. 
\begin{equation}
\small
    \text{AS} = \frac{1}{4} (\text{RANK}_{\text{PCC}} + \text{RANK}_{\text{SSIM}} + \text{RANK}_{\text{RMSE}} + \text{RANK}_{\text{JS}})
    \label{eq-as}
\end{equation}
where the average PCC/SSIM and RMSE/JS of all deconvolution methods are ranked in ascending and descending order, respectively, to obtain \(\text{RANK}_{\text{PCC}}\), \(\text{RANK}_{\text{SSIM}}\), \(\text{RANK}_{\text{RMSE}}\), and \(\text{RANK}_{\text{JS}}\).

\begin{table*}[htbp]%
\centering
\caption{Ablation experiments on the 32 simulated datasets in terms of the AS metric. AL: Adversarial Learning. mask: Reconstruction loss computed only for the masked points. Full\_rec: Full reconstruction loss. 
}\label{tab2}
\begin{adjustbox}{width=1\textwidth}
\renewcommand{\arraystretch}{0.9}  
\fontsize{9}{10}\selectfont        
\begin{tabular}{l|cccccccc}
\hline
\multicolumn{1}{l|}{\textbf{AS(↑)}}              & Dataset1             & Dataset2             & Dataset3             & Dataset4             & Dataset5             & Dataset6             & dataset7             & dataset8             \\ \hline
\multicolumn{1}{l|}{MACD (w/o mask, w/o AL)} & 0.638±0.054          & 0.563±0.042          & 0.538±0.082          & 0.363±0.021          & 0.425±0.018          & 0.438±0.052          & 0.513±0.036          & 0.413±0.036          \\
MACD (w/ mask,w/o AL)                         & 0.425±0.049          & 0.813±0.017          & 0.313±0.051          & 0.425±0.014          & 0.475±0.031          & 0.625±0.008          & 0.425±0.024          & 0.413±0.009          \\

MACD (w/o mask, w/ AL)                       & 0.700±0.008          & 0.363±0.021          & 0.725±0.018          & 0.800±0.007          & 0.525±0.011          & 0.650±0.035          & 0.638±0.026          & 0.800±0.012          \\

MACD (w/ Full\_rec, w/o AL)                    & 0.513±0.011          & 0.375±0.018          & 0.625±0.034          & 0.550±0.065          & 0.575±0.058          & 0.375±0.016          & 0.463±0.041          & 0.475±0.048          \\

\multicolumn{1}{l|}{MACD}                        & 0.725±0.068          & 0.888±0.007          & 0.800±0.015          & 0.863±0.024          & 0.998±0.001          & 0.913±0.011          & 0.963±0.006          & 0.900±0.015          \\ \hline
                                                & \multicolumn{1}{l}{} & \multicolumn{1}{l}{} & \multicolumn{1}{l}{} & \multicolumn{1}{l}{} & \multicolumn{1}{l}{} & \multicolumn{1}{l}{} & \multicolumn{1}{l}{} & \multicolumn{1}{l}{} \\ \hline
\multicolumn{1}{l|}{\textbf{AS(↑)}}              & Dataset9             & Dataset10            & Dataset11            & Dataset12            & Dataset13            & Dataset14            & Dataset15            & Dataset16            \\ \hline
\multicolumn{1}{l|}{MACD (w/o mask, w/o AL)} & 0.550±0.022          & 0.425±0.028          & 0.400±0.008          & 0.450±0.022          & 0.400±0.012          & 0.400±0.062          & 0.363±0.027          & 0.513±0.051          \\
MACD (w/ mask, w/o AL)                         & 0.625±0.016          & 0.375±0.038          & 0.388±0.012          & 0.588±0.037          & 0.763±0.009          & 0.650±0.007          & 0.425±0.001          & 0.400±0.028          \\

MACD (w/o  mask, w/ AL)                       & 0.700±0.008          & 0.825±0.004          & 0.775±0.011          & 0.663±0.052          & 0.638±0.027          & 0.588±0.006          & 0.750±0.028          & 0.563±0.004          \\
MACD (w/ Full\_rec, w/o AL)                    & 0.325±0.001          & 0.425±0.011          & 0.463±0.037          & 0.475±0.084          & 0.325±0.004          & 0.363±0.012          & 0.550±0.035          & 0.588±0.021          \\
\multicolumn{1}{l|}{MACD}                        & 0.800±0.035          & 0.950±0.005          & 0.975±0.003          & 0.825±0.038          & 0.875±0.008          & 1.000±0.000          & 0.913±0.004          & 0.938±0.006          \\ \hline
                                                & \multicolumn{1}{l}{} & \multicolumn{1}{l}{} & \multicolumn{1}{l}{} & \multicolumn{1}{l}{} & \multicolumn{1}{l}{} & \multicolumn{1}{l}{} & \multicolumn{1}{l}{} & \multicolumn{1}{l}{} \\ \hline
\multicolumn{1}{l|}{\textbf{AS(↑)}}              & Dataset17            & Dataset18            & Dataset19            & Dataset20            & Dataset21            & Dataset22            & Dataset23            & Dataset24            \\ \hline
\multicolumn{1}{l|}{MACD (w/o  mask, w/o  AL)} & 0.425±0.024          & 0.750±0.032          & 0.375±0.018          & 0.363±0.036          & 0.588±0.007          & 0.525±0.041          & 0.450±0.015          & 0.425±0.041          \\
MACD (w/ mask,w/o AL)                         & 0.488±0.074          & 0.400±0.032          & 0.575±0.091          & 0.325±0.023          & 0.650±0.002          & 0.488±0.024          & 0.525±0.034          & 0.888±0.021          \\

MACD (w/o  mask, w/ AL)                       & 0.638±0.012          & 0.438±0.036          & 0.650±0.055          & 0.775±0.003          & 0.650±0.045          & 0.763±0.026          & 0.500±0.012          & 0.338±0.002          \\
MACD (w/ Full\_rec,w/o AL)                    & 0.550±0.052          & 0.488±0.046          & 0.650±0.062          & 0.638±0.037          & 0.238±0.006          & 0.400±0.073          & 0.525±0.034          & 0.563±0.037          \\
\multicolumn{1}{l|}{MACD}                        & 0.900±0.015          & 0.925±0.003          & 0.750±0.138          & 0.900±0.028          & 0.875±0.014          & 0.825±0.038          & 0.998±0.001          & 0.788±0.022          \\ \hline
                                                & \multicolumn{1}{l}{} & \multicolumn{1}{l}{} & \multicolumn{1}{l}{} & \multicolumn{1}{l}{} & \multicolumn{1}{l}{} & \multicolumn{1}{l}{} & \multicolumn{1}{l}{} & \multicolumn{1}{l}{} \\ \hline
\multicolumn{1}{l|}{\textbf{AS(↑)}}              & Dataset25            & Dataset26            & Dataset27            & Dataset28            & Dataset29            & Dataset30            & Dataset31            & Dataset32            \\ \hline
\multicolumn{1}{l|}{MACD (w/o  mask, w/o  AL)} & 0.463±0.026          & 0.413±0.057          & 0.488±0.002          & 0.388±0.022          & 0.613±0.036          & 0.550±0.047          & 0.588±0.052          & 0.538±0.056          \\
MACD (w/ mask,w/o AL)                         & 0.738±0.017          & 0.300±0.008          & 0.513±0.022          & 0.500±0.028          & 0.800±0.028          & 0.638±0.029          & 0.425±0.043          & 0.313±0.011          \\

MACD (w/o  mask, w/ AL)                       & 0.613±0.026          & 0.763±0.027          & 0.750±0.052          & 0.888±0.011          & 0.488±0.022          & 0.475±0.008          & 0.663±0.012          & 0.525±0.043          \\

MACD (w/ Full\_rec,w/o AL)                    & 0.275±0.009          & 0.613±0.042          & 0.325±0.004          & 0.450±0.042          & 0.400±0.022          & 0.513±0.012          & 0.363±0.021          & 0.738±0.021          \\

\multicolumn{1}{l|}{MACD}                        & 0.913±0.002          & 0.913±0.011          & 0.925±0.004          & 0.775±0.069          & 0.700±0.105          & 0.825±0.014          & 0.963±0.002          & 0.888±0.037          \\ \hline
\end{tabular}
\end{adjustbox}
\end{table*}

\section{EXPERIMENTAL RESULTS}\label{result}
\subsection{Dataset description}\label{data}
This study utilized 32 simulated datasets from the benchmark study \cite{Li2022} and two real datasets from distinct tissues: one from mouse lymph node tissue \cite{DestVI} and the other from human developing heart tissue \cite{heart} (\autoref{tab1}). The scRNA-seq data in the 32 simulated datasets were derived from real tissues, with the corresponding ST data being simulated. The ST data (10x Visium) for the mouse lymph node include 1,092 spots, while the scRNA-seq data (10x Chromium) encompass 14,989 cells classified into 15 cell types. For the human developing heart tissue, the ST (ST) data comprise 210 spots, and the scRNA-seq data (10x Chromium) contain 3,777 single cells, identifying 15 distinct cell types.

\textbf{ST and scRNA-seq data preprocessing.} 
We employed tools such as Seurat \cite{Seurat} and Scanpy \cite{Scanpy} to perform cell clustering on the scRNA-seq data, identify marker genes for each cell type, and retain the top 200 marker genes per cell type. For the ST data, we retained only the genes that intersected between the ST data and the preprocessed scRNA-seq data. Detailed descriptions of each dataset are provided in \autoref{tab1}. 

\textbf{Simulated ST data.} 
The process of generating pseudo-spot based Simulated ST data from scRNA-seq data is referenced in \cite{spoint}. Simply, based on these single cells from the referenced scRNA-seq data, we generate pseudo-spots through a simulation process to obtain pseudo-spot based \textbf{Simulated ST data}.

\subsection{Baseline methods}\label{Comparison with Other Prediction Methods}
In this study, we selected six representative state-of-the-art
methods:
\begin{itemize}
    \item Spoint \cite{spoint}: Spoint utilizes deep learning for feature extraction in spatial transcriptomics data and applies PCA for dimensionality reduction, enhancing cell type deconvolution accuracy.
    \item Tangram \cite{Tangram}: Tangram employs deep learning to map scRNA-seq data onto ST data, predicting the spatial distribution of cell types.
    \item Cell2location \cite{Cell2location}: Cell2location performs cell-type deconvolution of ST data using a hierarchical Bayesian framework to achieve accurate results.
    \item DestVI \cite{DestVI}: DestVI utilizes variational inference and latent variable models to estimate the proportions of different cell types.
    \item scpDeconv \cite{scp}: scpDeconv combines scRNA-seq data with proteomic data using a domain adversarial autoencoder to improve the precision of identifying cell types.
    \item Stereoscope \cite{Stereoscope}: Stereoscope integrates scRNA-seq and ST data to spatially map cell types by leveraging probabilistic modeling techniques.
\end{itemize}

\subsection{Implementation Details}\label{data}
All baseline models were implemented with the default parameters specified in their original papers. Experiments were conducted on an NVIDIA RTX 3090 GPU using PyTorch (version 1.12.1) and Python 3.9. The training process was set for 200 epochs, with a batch size of 2048 , a learning rate of 0.01 and $\rho$ = 0.3.

We evaluated the model performance with different values of $\lambda$ in \autoref{allloss}.
When $\lambda$ = 0.1, the average AS for the 32 simulated datasets is 0.61; when $\lambda$ = 0.5, it is 0.75; and when $\lambda$ = 0.9, it is 0.63. Therefore, $\lambda$ = 0.5 was used for all subsequent experiments.

\subsection{Performance of MACD in the 32 simulated datasets}\label{Performance of simulated datasets}
Due to the lack of single-cell resolution in real ST data, it is impossible to precisely quantify the accuracy of our inferred cell type compositions. To assess the performance of MACD in inferring cell type compositions, we used 32 simulated datasets from the benchmark study \cite{Li2022}. These datasets are designed to closely mimic real ST data scenarios, providing high reliability for our evaluation. We compared MACD with six state-of-the-art cell type deconvolution methods in terms of performance. 

Our experimental results showed that among the six deconvolution methods, MACD achieved the highest average PCC/SSIM values (0.941/0.929) and the lowest average RMSE/JS values (0.043/0.162) (See~\autoref{fig2}A). Furthermore, we evaluated overall performance using the AS. MACD achieved a significantly higher average AS (0.93) compared to other methods (AS = 0.26–0.76) (See~\autoref{fig2}B). Overall, these results demonstrate that MACD offers substantial advantages in cell type deconvolution, showing superior performance in both similarity assessment and error measurement. Through comprehensive testing on simulated datasets, MACD has proven its robust capability in predicting complex cell type compositions.

\subsection{Performance of MACD in the MLN dataset}\label{Performance of  murine lymph}
To evaluate  the effectiveness of MACD on real tissue samples, we applied it to a mouse lymph node (MLN) dataset \cite{DestVI}. We analyzed mature B cells and CD8 T cells from MLN tissue, where mature B cells produce antibodies and CD8 T cells target infected or abnormal cells.
Using the differential analysis tools provided by Scanpy \cite{Scanpy},  
we identified key marker genes for cell type characterization. Specifically, \textit{Cd79a} was selected as a marker for mature B cells, and \textit{Klk8} was chosen for CD8 T cells. The predicted results of mature B cells and CD8 T cells by MACD closely match the marker gene expression levels (\autoref{fig3}A and \autoref{fig3}B). We performed quantitative evaluation using these marker genes as benchmarks. The PCC between the predicted mature B cells and the marker gene \textit{Cd79a} was 0.8049, with a JS of 0.0612 (\autoref{fig3}C). For CD8 T cells, the PCC with the marker gene Klk8 was 0.3615, and the JS was 0.2206 (\autoref{fig3}D).
Compared to six state-of-the-art deconvolution methods,MACD demonstrated superior performance in predicting cell type proportions, achieving lower error rates.

\subsection{Performance of MACD in the HDH dataset}\label{Performance of heart}
To further evaluate MACD’s performance on different tissue types, we applied it to the human developing heart (HDH) dataset \cite{heart}. We assessed atrial cardiomyocytes using \textit{MYH6} as the marker gene, which are crucial for efficiently transferring blood from the atria to the ventricles and maintaining normal heart function. MACD’s predictions for atrial cardiomyocyte distribution were highly consistent with \textit{MYH6} expression (\autoref{fig4}A), achieving the highest PCC of 0.5721 and the lowest JS of 0.0657 (\autoref{fig4}C). These results highlight MACD's effectiveness in deconvoluting rare cell types.

Subsequently, we assessed regional-enriched ventricular cardiomyocytes using \textit{MYH7} as the marker gene. Ventricular cardiomyocytes are located in the heart's ventricles and are critical for systemic blood circulation. The results indicated that MACD’s predicted distribution of ventricular cardiomyocytes closely matched the marker gene (\autoref{fig4}B). In the quantitative assessment (\autoref{fig4}D), MACD achieved the highest PCC of 0.8542 and the lowest JS of 0.0301, confirming its ability to accurately predict the proportions and spatial distribution of ventricular cardiomyocytes. Overall, these results confirm that MACD effectively performs cell type deconvolution in the HDH dataset.

\subsection{Ablation studies}\label{Model robustness}
To evaluate the contributions of the masking and adversarial modules to the MACD model, we conducted a series of ablation experiments (\autoref{tab2}). We specifically aimed to determine if these modules enhance model performance and whether reconstruction loss should be computed on the masked region or the entire ST data. The experiments included a baseline model with only the encoder and predictor, a model with a masking module (reconstruction loss computed only for the masked region), a model with full reconstruction loss, and a model with an adversarial module. The ablation experiments confirm the critical role of the masking and adversarial modules in enhancing the accuracy of cell type proportion predictions.

\section{CONCLUSIONS}\label{CONCLUSIONS AND DISCUSSION}

To address the challenge of single-cell resolution in spatial transcriptomics (ST) technologies, we present MACD, a masked adversarial neural network designed for precise cell type deconvolution. MACD leverages a masked autoencoder (MAE) to effectively learn latent features from real ST data during encoding and decoding. By employing adversarial learning, MACD aligns real and simulated ST data in a unified latent space, thereby reducing discrepancies. Furthermore, MACD infers cell type compositions through supervised learning on labeled simulated ST data. Evaluations on 32 simulated datasets and 2 real datasets show that MACD significantly outperforms six state-of-the-art deconvolution methods, demonstrating superior performance across these datasets. Ablation studies highlight the critical role of each component in the MACD framework. 

\section*{Acknowledgment}
The work was supported in part by the National Natural Science Foundation of China (62262069, 62062067).

\balance
\small
\bibliography{references.bib} 
\bibliographystyle{IEEEtran}  
\end{document}